\documentclass[final]{cvpr}

\usepackage{times}
\usepackage{epsfig}
\usepackage{graphicx}
\usepackage{amsmath}
\usepackage{amssymb}
\usepackage[utf8]{inputenc}
\usepackage[T1,T5]{fontenc}

\usepackage[dvipsnames]{xcolor}
\usepackage[numbers,sort]{natbib}
\usepackage[hyphens]{url} %
\usepackage{tabularx}
\usepackage{mathtools}
\usepackage{booktabs}
\usepackage{ulem}
\usepackage{enumitem}%
\usepackage{verbatim}
\usepackage{subcaption}
\usepackage{amsfonts} %
\DeclareMathSymbol{\shortminus}{\mathbin}{AMSa}{"39}

\usepackage{comment}
\usepackage{soul}
\definecolor{cadmiumgreen}{rgb}{0.0, 0.42, 0.24}

\DeclareMathSymbol{\shortminus}{\mathbin}{AMSa}{"39}
\newcommand{\fig}[1]{Fig~\ref{fig:#1}}
\newcommand{\sect}[1]{Sect~\ref{sect:#1}}

\newcommand{\eq}[1]{Eq. (\ref{eq:#1})}

\newcommand\RGB{\textcolor{black}{R}\textcolor{black}{G}\textcolor{black}{B }}

\usepackage{xspace}

\usepackage[pagebackref=true,breaklinks=true,colorlinks,bookmarks=false]{hyperref}
\newcommand{\MethodName}{FaceDet3D\xspace}

\def\mbf#1{\mathbf{#1}}
\def\mbb#1{\mathbb{#1}}
\def\mcal#1{\mathcal{#1}}

\def\mtxtlog{\text{log }}
\usepackage{environ}
\NewEnviron{smequation}{
    \small
    \begin{equation}
        \BODY
    \end{equation}
}
\NewEnviron{smequation*}{%
    \begin{equation*}
    \scalebox{.93}{$\BODY$}
    \end{equation*}
}

\begin{document}

\title{\MethodName: Facial Expressions with 3D Geometric Detail Prediction}
\author{ShahRukh Athar$^{1}$
\and
Albert Pumarola$^{2}$
\and 
Francesc Moreno-Noguer$^{2}$
\and
Dimitris Samaras$^{1}$
\\~\\
${ }^{1}$Stony Brook University\\
${ }^{2}$Institut de Rob\`{o}tica i Inform\`{a}tica Industrial, CSIC-UPC
}

\maketitle

\begin{abstract}
Facial Expressions induce a variety of high-level details on the 3D face geometry. For example, a smile causes the wrinkling of cheeks or the formation of dimples, while being angry often causes wrinkling of the forehead. Morphable Models (3DMMs) of the human face fail to capture such fine details in their PCA-based representations  and consequently cannot generate such details when  used to edit expressions. In this work, we introduce \MethodName, a first-of-its-kind method that generates - from a single image - geometric facial details that are consistent with any desired target expression.  The facial details are represented as a vertex displacement map and used then by a Neural Renderer to photo-realistically render novel images of any single image in any desired expression and view. The project website can be found \href{http://shahrukhathar.github.io/2020/12/14/FaceDet3D.html}{here}.

\end{abstract}
\vspace{-0.5cm}
\section{Introduction}
Modelling the geometry of the human face continues to attract great interest
in the computer vision and computer graphics communities. Strong PCA-based priors make 3D morphable models (3DMMs) \cite{blanz1999morphable} robust, 
but at the same time over-regularize them,  so they fail to capture fine facial details,  such as the wrinkles on the forehead when the eyebrows are raised or  bumps on the cheeks when one smiles. Additionally, the lack of diversity in the texture space of most available 3DMMs make it very hard to generate realistic renderings that capture the large variations of color and texture we observe in human faces. Recent methods \cite{tewari2019fml, tewari2017self, tran2018nonlinear, tran2019towards, reda, booth20173d, dou2017end, jackson2017large, kim2018inversefacenet, Genova_2018_CVPR} address this by learning richer shape and expression spaces using a variety of data modalities such as in-the-wild images \cite{tewari2017self, tran2018nonlinear, tran2019towards, reda} and videos \cite{tewari2019fml}. However, despite using  more expressive shape and expression spaces, these models still fail to capture fine details in geometry.

\begin{figure}[t!]
    \hspace{-0.7cm}\includegraphics[width=1.15\linewidth]{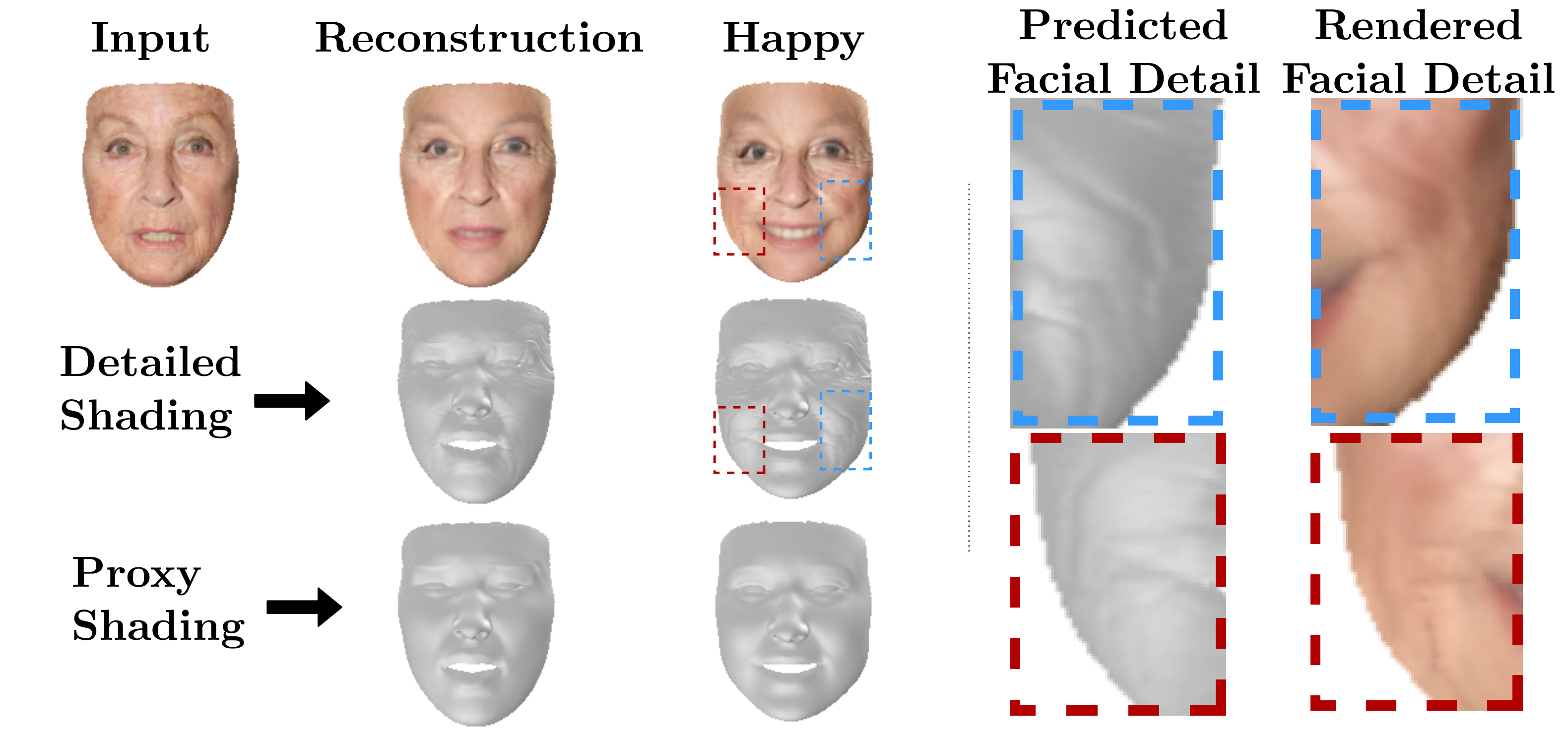}
    \caption{\small{\textbf{Facial detail prediction and rendering.} Given a single input image and a target expression (in this case `Happy'), our method predicts facial geometric details consistent with the target expression and renders realistic images. \textit{(Electronic zoom recommended)}}\vspace{-0.7cm}}
    \label{fig:teaser}
\end{figure}

Recent methods that accurately estimate facial geometric details from single images \cite{FDS,extreme3D}, while being unable to predict novel details under expression change, have nonetheless enabled the large scale annotation of unpaired image data with facial geometric details. Thus, it is now possible to train facial detail prediction methods using unsupervised adversarial losses \cite{StarGAN2018, pumarola2020ganimation}. Similarly, Neural Rendering \cite{dnr} has made it possible to render 3D geometries with photo-realistic detail via the use of high-dimensional Neural Texture Maps (NTMs). However, NTMs are able to store fine details of the output image, causing the rendered details to be completely independent of the geometric details. They do not change even if the geometric details do, making neural rendering unsuitable for rendering facial geometric details.  
In this paper we introduce \MethodName, which is, to the best of our knowledge, the first generative model that predicts facial geometric details for any target expression and renders them realistically to the image space.
The model is made up of two components: 1) A detail prediction network that infers plausible geometric facial details of a person as their expression changes, given a \textit{single} image of that person. 
2) A rendering network that overcomes the aforementioned shortcoming of neural rendering and explicitly conditions the rendered facial details on the \textit{details} of the 3D face geometry. This conditioning is achieved through the use of the novel \textit{Augmented Wrinkle Loss} and the \textit{Detailed Shading Loss} during training.

 Our method is trained using only a large scale in-the-wild image dataset and a much smaller video dataset captured in controlled conditions, without any 3D data as supervision for the target expression geometry. An exhaustive evaluation shows that once trained, our method is able to generate plausible facial details for any desired anatomically consistent facial expression and render it realistically in any desired view direction.

\section{Related Work}
We next describe the most related methods in facial geometry estimation, geometric facial detail estimation and facial expression editing and reanimation.
\vspace{-0.4cm}
\paragraph{Facial Geometry Estimation} Traditional facial geometry estimation methods fit statistical 3D face models to a given image, going back to the original 3D Face Morphable Model \cite{blanz1999morphable}. However,  the PCA-based prior used by 3DMMs is not flexible enough to represent fine facial geometric details such as wrinkles and skin bumps. Recent methods \cite{tewari2019fml, tewari2017self, tran2018nonlinear, tran2019towards, reda, booth20173d, dou2017end, jackson2017large, kim2018inversefacenet, Genova_2018_CVPR} leverage the power of deep-learning and large-scale image and video datasets  to regress parameters that generate realistic 3D reconstructions or learn  complex representations  for face shape, expression and texture. These methods generate  more realistic results than traditional 3DMM fitting, but  still cannot capture fine facial details. In our work we use the Basel Face Model \cite{gerig2018morphable} as the underlying geometry on top of which we predict  geometric details.%
\vspace{-0.4cm}
\paragraph{Geometric Facial Details Estimation.}  Over the past few years there has been significant  improvement in the realism of 3D face geometries estimated from single images. In \cite{sela2017unrestricted}, the  regressed correspondence and depth maps are registered onto a template mesh, which   is further refined to generate the detailed facial geometry. In \cite{extreme3D}   facial details are modelled with bump maps on top of a 3DMM base. %
Similarly, in Facial Details Synthesis (FDS)\cite{FDS},   details are represented as vertex displacements of an underlying 3DMM, trained using a combination of 3D data and in-the-wild images. These methods, however, can only estimate the facial details of the expression manifested in an image, but cannot predict novel facial details for a different expression, which is the motivation of our method. 

\vspace{-0.4cm}
\paragraph{Facial Expression Editing.} The success of image-to-image translation networks \cite{pix2pix2016}, adversarial training \cite{goodfellow2014generative} and cycle-consistency losses \cite{CycleGAN2017} have led to the development of expression editing methods that  use large-scale in-the-wild training datasets. In \cite{NeuralFace2017},  an unsupervised face model disentangles the input face into albedo, normals and shading. Expressions are then edited by traversals in the disentangled latent space. In \cite{StarGAN2018},   expressions are edited via adversarial losses coupled with cycle consistency. In \cite{pumarola2020ganimation},  a network  edits input images to target expressions,  represented as Action Units \cite{FACS}. \cite{athar2020self} extends this work by  explicitly modeling skin motion followed by  texture hallucination. %
In \cite{starganv2}, template images are used for editing. While these methods give photo-realistic results, they are restricted to the 2D Image space and  cannot be used to manipulate 3D viewpoint. 
\vspace{-0.4cm}
\paragraph{Facial Reanimation Methods.} Facial renanimation methods use driving parameters from a source video to reanimate a source image or  video.  \cite{Kim2018DeepVP}, performs the reanmatiion using the parameters of a 3DMM extracted from the source video. In \cite{dnr}, Neural Rendering is used to reanimate videos using expression parameters extracted from a source video. However, this renderer is  needs to be retrained for each target video. In \cite{Thies2020NeuralVP}, audio parameters drive the target video. In \cite{zakharov2019few},  facial landmarks from a source video  are used to animate single images. While all these methods generate realistic re-animations, they do not model any facial geometric detail. More closely related to our work is \cite{Nagano2018paGANRA}, which regresses a set of FACS \cite{FACS} based textures from single images and are able to generate realistic renderings in any desired expression. Nevertheless, they do not predict any facial geometric detail but instead generate the details in the texture space.
\section{\MethodName}
\subsection{Problem Formulation}
A face image \(I_{\mbf{x}} \in \mbb{R}^{H \times W \times 3}\) has expression \(\mbf{x}\) represented by Action Units \cite{FACS}. Its shape and expression parameters in the Basel Face Model (BFM) \cite{gerig2018morphable} space are \(\{\alpha_{s}, \alpha_{e}\}\). A vertex displacement UV map \(\mcal{D}(I_{\mbf{x}}) \in \mbb{R}^{H_{\mcal{D}}\times W_\mcal{D} \times 3}  \) encodes the facial details of the person shown in \(I_{\mbf{x}}\) with the expression \(\mbf{x}\). We want to: (1) Generate facial details, represented by a vertex displacement map, \(\mcal{D}(I_{\mbf{y}})\) as the expression of that person changes to \(\mbf{y}\); (2) Render an image of the geometry \textit{with} these details. %

We use  FDS \cite{FDS} to extract the texture map, initial detail map and geometry of the input image \(I_{\mbf{x}}\):
\begin{smequation}
    \mcal{T}(I_{\mbf{x}}), \mcal{D}(I_{\mbf{x}}), \alpha_{s}, \alpha_{e} \leftarrow FDS(I_{\mbf{x}})
\end{smequation}
where, \(\mcal{T}(I_{\mbf{x}}) \in \mbb{R}^{H_{\mcal{T}}\times W_\mcal{T} \times 3}\) is the texture map.

Next, we use the \textbf{detail prediction network}, \(DetP(\cdot)\), to predict the plausible facial detail map of the person in \(I_{\mbf{x}}\) for expression \(\mbf{y}\) as:
\begin{smequation}
    \mcal{\widetilde{D}}(I_{\mbf{y}}) = DetP(\mcal{D}(I_{\mbf{x}}), \mbf{x}, \mbf{y}, \hat{\alpha_{e}}, Age(I_{\mbf{x}}), FaceID(I_{\mbf{x}}))
    \label{eq:detpred_f}
\end{smequation}
 where, \(\mbf{y}\), \(\hat{\alpha_{e}}\) are the target AUs and expression parameters, \(Age(I_{\mbf{x}})\) are  features extracted from an age prediction network and \(FaceID(I_{\mbf{x}})\) is the facial embedding of \(I_{\bf{x}}\) extracted using \cite{schroff2015facenet}.
 Note, we \textit{do not have access} to the ground truth image \(I_{\mbf{y}}\), thus we \textit{predict a plausible detail map} of \(I_{\mbf{y}}\) (i.e \(\mcal{\widetilde{D}}(I_{y})\)) using \(DetP\). Once we have \(\mcal{\widetilde{D}}(I_{y})\), we use it to displace the vertices along their normal direction giving us the detailed geometry. Finally, we render this detailed face geometry using a \textbf{rendering network} \(R(\cdot)\):
 \begin{smequation}
     \tilde{I}_{\bf{y}} = R(\mcal{T}(I_{\mbf{x}}), \mcal{\widetilde{D}}({I}_{\mbf{y}}), \alpha_{s}, \hat{\alpha_{e}}, \mbf{y}, c, l, \gamma)
     \label{eq:refnet_f}
 \end{smequation}
where, \(c\) are the desired camera and view parameters, \(\gamma\) is the albedo PCA-space parameters of BFM\cite{gerig2018morphable}, and \(l\) are the lighting parameters.

\subsection{Detail Prediction}
\begin{figure*}[t!]
    \centering
    \includegraphics[width=0.72\linewidth]{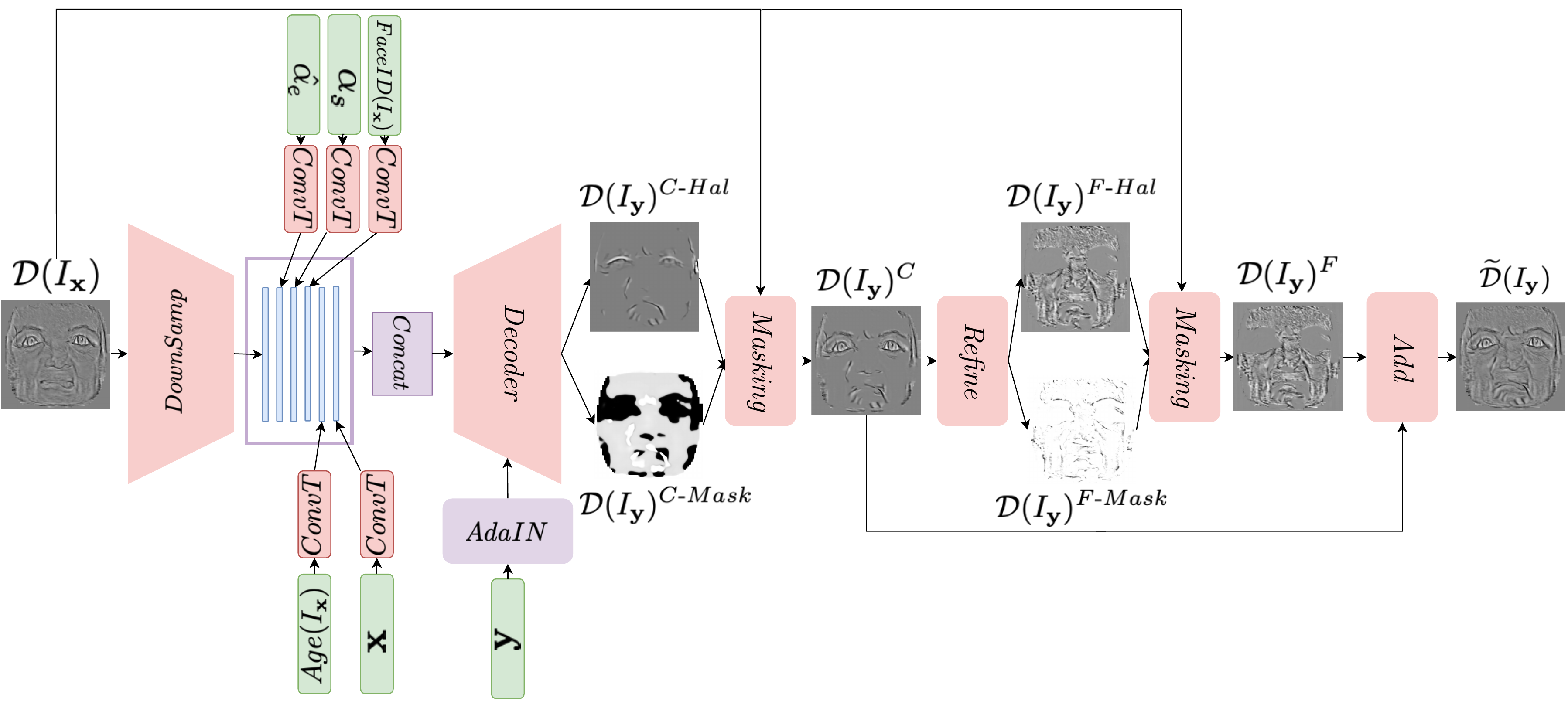}
    \caption{\small{\textbf{Detail Prediction Network.} \(DetP\) first extracts from its various inputs high-dimensional features and concatenates them. This concatenated feature map is then used to predict a coarse detail map \(\mcal{D}(I_{\mbf{y}})^{C}\). The coarse detail map is then refined via a refinement module to predict a fine detail map \(\mcal{D}(I_{\mbf{y}})^{F}\) which is then added to the coarse detail map,\(\mcal{D}(I_{\mbf{y}})^{C}\), to give \(\widetilde{\mcal{D}}(I_{\mbf{y}})\).}\vspace{-0.3cm}}
    \label{fig:detp_arch}
\end{figure*}
 \(DetP\) takes the input detail map, \(\mcal{D}(I_{\mbf{x}})\), the target expression parameters \(\hat{\alpha_{e}}\), the shape parameters \(\alpha_{shape}\), the face embedding \(FaceID(I_{\mbf{x}})\), the action unit \(\mbf{x}\), and the age features  \(Age(I_{\mbf{x}})\) and extracts features from each of them. There features are concatenated in the channel dimension and passed though the \(Decoder\) which receives the target action unit \(\mbf{y}\) via Adaptive Instance Normalization \cite{adain} layers. The \(Decoder\) gives as output a hallucination \(\mcal{D}(I_{\mbf{y}})^{C\text{-}Hal}\) and a mask \(\mcal{D}(I_{\mbf{y}})^{C\text{-}Mask}\) which are combined together to give \(\mcal{D}(I_{\mbf{y}})^{C}\):
\begin{smequation}
    \begin{split}
        \mcal{D}(I_{\mbf{y}})^{C} &= \mcal{D}(I_{\mbf{y}})^{C\text{-}Mask} \odot \mcal{D}(I_{\mbf{y}})^{C\text{-}Hal} \\
        &+ (1 \shortminus \mcal{D}(I_{\mbf{y}})^{C\text{-}Mask}) \odot \mcal{D}(I_{\mbf{x}})
    \end{split}
\end{smequation}
\(\mcal{D}(I_{\mbf{y}})^{C}\) then goes through a refinement module that generates \(\mcal{D}(I_{\mbf{y}})^{F\text{-}Hal}\) and a mask \(\mcal{D}(I_{\mbf{y}})^{F\text{-}Mask}\) which are combined to give \(\mcal{D}(I_{\mbf{y}})^{F}\):
\begin{smequation}
    \begin{split}
        \mcal{D}(I_{\mbf{y}})^{F} &= \mcal{D}(I_{\mbf{y}})^{F\text{-}Mask} \odot \mcal{D}(I_{\mbf{y}})^{F\text{-}Hal} \\
        &+ (1 \shortminus \mcal{D}(I_{\mbf{y}})^{F\text{-}Mask}) \odot \mcal{D}(I_{\mbf{x}})
    \end{split}
\end{smequation}
Finally, \(\widetilde{\mcal{D}}(I_{\mbf{y}}) = \mcal{D}(I_{\mbf{y}})^{C} + \mcal{D}(I_{\mbf{y}})^{F}\). %
With this two-stage approach \(\mcal{D}(I_{\mbf{y}})^{C}\) is able to capture coarse-scale detail changes  while \(\mcal{D}(I_{\mbf{y}})^{F}\) captures the finer scale changes. The masking mechanism used in both stages ensures the preservation of  the details that are not meant to be changed with the expression. %
\vspace{-0.3cm}
\subsubsection{Training Losses}
Due to the lack of a large scale in-the-wild dataset of paired data with expression change or 3D data, we cannot directly perform a regression using ground-truth geometric facial details, \(\mcal{D}(I_{\mbf{y}})\) of the image \(I_{\mbf{y}}\). Therefore, we instead use an adversarial training paradigm along with cycle-consistency losses similar to \cite{athar2020self, pumarola2020ganimation} to predict the plausible facial geometric details \(\mcal{\widetilde{D}}(I_{y})\) and to ensure the prediction's fidelity to the target expression and input features. In order to speed up convergence, we weakly supervise the adversarial training using randomly sampled frames from videos of the MUG \cite{MUG} and the ADFES datasets \cite{ADFES}, ensuring that the sampling is sparse enough such that there is significant change in expression with frames sampled from each video. 
\vspace{-0.6cm}
\paragraph{Expression Adversarial Loss.} In order to ensure the predicted facial geometric details, \(\mcal{\widetilde{D}}(I_{\mbf{y}})\), are consistent with the target expression \(\{\mbf{y}, \hat{\alpha}_{e}\}\), we use an expression discriminator \(\textrm{D}_{\textrm{Exp}}\). Given  \(\mcal{D}(I_{\mbf{x}})\) of some image \(I_{\mbf{x}}\) manifesting expression \(\{\mbf{x}, \alpha_{e}\}\), \(\textrm{D}_{\textrm{Exp}}\), outputs the following
\begin{smequation}
    \textrm{D}_{\textrm{Exp}}(\mcal{D}(I_{x})) = \{r, \hat{\mbf{x}}, \hat{\alpha_{e}}\}
\end{smequation}
where \(r\) is a realism score and \(\hat{\mbf{x}} \text{ and } \hat{\alpha_{e}}\) are the predicted AUs and expression parameters, respectively. For brevity, we will use \(\textrm{D}_{\textrm{Exp}}(\mcal{D}(I_{\mbf{x}}))\) and \(\textrm{D}_{\textrm{Exp}}(\mcal{D})\) interchangeably. We use the Non-Saturating adversarial loss \cite{goodfellow2014generative} along with the \(R1\) gradient penalty \cite{mescheder2018training} to train \(\textrm{D}_{\textrm{Exp}}\). Specifically, let \(\mcal{P}_{\mcal{D}}\) be the distribution of real facial geometric detail maps, the adversarial loss for \(\textrm{D}_{\textrm{Exp}}\) can   be written as:
\begin{smequation}
\begin{split}
    \mcal{L}^{\textrm{D}_{\textrm{Exp}}}_{Adv} &= -\mathbb{E}_{\mcal{\widetilde{D}} \sim DetP(.)}\left[\mtxtlog \left( 1 \shortminus \textrm{D}_{\textrm{Exp}}^{r}(\mcal{\widetilde{D}})\right)\right]\\
                            &- \mathbb{E}_{\mcal{D} \sim \mcal{P}_{\mcal{D}}}\left[\mtxtlog \left(\textrm{D}_{\textrm{Exp}}^{r}(\mcal{D})\right)\right]\\
                            &+ \mathbb{E}_{\mcal{D} \sim \mcal{P}_{\mcal{D}}}\left[||\nabla \textrm{D}_{\textrm{Exp}}^{r}(\mcal{D})||_{2}^{2}\right]
\end{split}
\label{eq:adv_gen}
\end{smequation}
where \(\textrm{D}_{\textrm{Exp}}^{r}\) is the realism output head of \(\textrm{D}_{\textrm{Exp}}\). In addition, \(\textrm{D}_{\textrm{Exp}}\) is trained to minimize the error of the predicted AU and expression parameters
\begin{smequation}
    \begin{split}
        \mcal{L}^{\textrm{D}_{\textrm{Exp}}}_{Exp} &= \mathbb{E}_{\mcal{D} \sim \mcal{P}_{\mcal{D}}}\left[||[\textrm{D}_{\textrm{Exp}}^{\textrm{AU}}(\mcal{D}) \shortminus \mbf{x}||_{2}^{2}\right]\\
        & + \mathbb{E}_{\mcal{D} \sim \mcal{P}_{\mcal{D}}}\left[||[\textrm{D}_{\textrm{Exp}}^{\alpha_{e}}(\mcal{D}) - \alpha_{e}||_{2}^{2}\right]
    \end{split}
\end{smequation}
where \(\textrm{D}_{\textrm{Exp}}^{\textrm{AU}} \text{ and }\textrm{D}_{\textrm{Exp}}^{\alpha_{e}} \) are the AU and expression parameter output head of \(\textrm{D}_{\textrm{Exp}}\) respectively. The Detail Prediction Network, \(DetP\) is trained to minimize the adversarial loss
\begin{smequation}
    \begin{split}
        \mcal{L}^{DetP}_{ExpAdv} &= \underset{I_{\mbf{x}}, \{\mbf{y}, \hat{\alpha}_{e}\} }{\shortminus\mathbb{E}}\mtxtlog \left(\textrm{D}_{\textrm{Exp}}^{r}(DetP(\cdot))\right)\\
    \end{split}
\end{smequation}
and the expression losses
\begin{smequation}
    \begin{split}
        \mcal{L}^{DetP}_{AU} &= \underset{I_{\mbf{x}}, \{\mbf{y}, \hat{\alpha}_{e}\} }{\mathbb{E}}||\textrm{D}_{\textrm{Exp}}^{\textrm{AU}}(DetP(\cdot)) \shortminus \mbf{y}||_{2}^{2}\\
    \end{split}
\end{smequation}
\begin{smequation}
    \begin{split}
        \mcal{L}^{DetP}_{\alpha_{e}} &= \underset{I_{\mbf{x}}, \{\mbf{y}, \hat{\alpha}_{e}\} }{\mathbb{E}}||\textrm{D}_{\textrm{Exp}}^{\alpha_{e}}(DetP(\cdot)) \shortminus \hat{\alpha}_{e}||_{2}^{2}\\
    \end{split}
\end{smequation}
where, \(DetP(\cdot)\) is to be read as in \eq{detpred_f} and \(\mbf{y}\) and \(\hat{\alpha}_{e}\) are the target AU and expression parameters respectively.
\vspace{-0.3cm}
\paragraph{FaceID Loss.} The FaceID loss ensures that facial details characteristic of the subject's identity,  that are invariant to expression change, are preserved in the predicted details. A face embedding detection network, \(\textrm{D}_{\textrm{Face}}\), is trained to predict the \(FaceNet\) \cite{schroff2015facenet} features of an image \(I_{\mbf{x}}\) from its detail map \(\mcal{D}(I_{\mbf{x}})\) as follows
\begin{smequation}
    \begin{split}
        \mcal{L}^{\textrm{D}_{\textrm{Face}}}_{ID} &= \mathbb{E}_{\mcal{D} \sim \mcal{P}_{\mcal{D}}}||\textrm{D}_{\textrm{Face}}^{\textrm{ID}}(\mcal{D}(I_{\mbf{x}})) \shortminus FaceNet(I_{\mbf{x}})||_{2}^{2}
    \end{split}
\end{smequation}

\noindent where \(\textrm{D}_{\textrm{Face}}^{\textrm{ID}}\) is the face embedding output head of \(\textrm{D}_{\textrm{Face}}\). Additionally, \(\textrm{D}_{\textrm{Face}}\) is also trained with an adversarial loss with the same loss as shown in \eq{adv_gen}. The detail prediction network, \(DetP\) is trained to minimize both the FaceID loss and the adversarial loss
\begin{smequation}
    \begin{split}
        \mcal{L}^{DetP}_{ID} &= \underset{I_{\mbf{x}}, \{\mbf{y}, \hat{\alpha}_{e}\}}{\mathbb{E}}  ||\textrm{D}_{\textrm{Face}}^{\textrm{ID}}(DetP(\cdot)) \shortminus FaceNet(I_{\mbf{x}})||_{2}^{2}\\
    \end{split}
\end{smequation}
\begin{smequation}
    \begin{split}
        \mcal{L}^{DetP}_{IDAdv} &= \underset{I_{\mbf{x}}, \{\mbf{y}, \hat{\alpha}_{e}\} }{\shortminus\mathbb{E}}\mtxtlog \left(\textrm{D}_{\textrm{Face}}^{\textrm{r}}(DetP(.)\right)\\
    \end{split}
\end{smequation}
where \(\textrm{D}_{\textrm{Face}}^{\textrm{r}}\) is the realism output head of \(\textrm{D}_{\textrm{Face}}\).
\vspace{-0.3cm}
\paragraph{Age Loss.} The Age Loss ensures that the facial details predicted are consistent with the subject's age  by ensuring the age feature embedding of the predicted detail map matches that of the input details. An age feature prediction network, \(\textrm{D}_{\textrm{Age}}\) is trained to predict the features of an image \(I_{\mbf{x}}\) extracted using a pre-trained age prediction network, \(AgeNet\), from its  detail map \(\mcal{D}(I_{\mbf{x}})\) as follows
\begin{smequation}
    \begin{split}
        \mcal{L}^{\textrm{D}_{\textrm{Age}}}_{Age} &= \mathbb{E}_{\mcal{D} \sim \mcal{P}_{\mcal{D}}}||\textrm{D}_{\textrm{Age}}^{\textrm{Age}}(\mcal{D}(I_{\mbf{x}})) \shortminus AgeNet(I_{\mbf{x}})||_{2}^{2}
    \end{split}
\end{smequation}
where \(\textrm{D}_{\textrm{Age}}^{\textrm{Age}}\) is the age feature output head of \(\textrm{D}_{\textrm{Age}}\). Additionally, \(\textrm{D}_{\textrm{Age}}\) is also trained with an adversarial loss of the same type as shown in \eq{adv_gen}. The detail prediction network, \(DetP\) is trained to minimize both the Age loss and the adversarial loss
\begin{smequation}
    \begin{split}
        \mcal{L}^{DetP}_{Age} &= \underset{I_{\mbf{x}}, \{\mbf{y}, \hat{\alpha}_{e}\}}{\mathbb{E}}  ||\textrm{D}_{\textrm{Age}}^{\textrm{Age}}(DetP(.)) \shortminus AgeNet(I_{\mbf{x}})||_{2}^{2}\\
    \end{split}
\end{smequation}
\begin{smequation}
    \begin{split}
        \mcal{L}^{DetP}_{AgeAdv} &= \underset{I_{\mbf{x}}, \{\mbf{y}, \hat{\alpha}_{e}\} }{-\mathbb{E}}\mtxtlog \left(\textrm{D}_{\textrm{Age}}^{\textrm{r}}(DetP(.)\right)\\
    \end{split}
\end{smequation}
where \(\textrm{D}_{\textrm{Age}}^{\textrm{r}}\) is the realism output head of \(\textrm{D}_{\textrm{Age}}\).
\vspace{-0.3cm}
\paragraph{Regression Loss.}
In order to speed up training, we use a small amount of video data from  MUG \cite{MUG} and ADFES \cite{ADFES} to directly regress the details map estimated by FDS \cite{FDS} as
\begin{smequation}
    \begin{split}
     &\hspace{-0.1cm}\mcal{\widetilde{D}}(I_{\bf{y}}^{k}) = DetP(\mcal{D}(I_{\bf{x}}^{m}), \mbf{x}, \mbf{y}, \hat{\alpha_{e}}, Age(I_{\bf{x}}^{m}), FaceID(I_{\bf{x}}^{m}))\\
    &\mcal{L}_{Regress}^{DetP} = LapLoss(\mcal{\widetilde{D}}(I_{\bf{y}}^{k}), \mcal{D}(I_{\bf{y}}^{k}))  
    \end{split}
\end{smequation}
where, \(\mcal{D}(I_{\bf{y}}^{k})\) and \(\mcal{D}(I_{\bf{x}}^{m})\) are the detail map of \(k\)-th frame \(I_{\bf{y}}^{k}\) and \(m\)-th frame \(I_{\bf{x}}^{m}\) respectively.
Training solely on video data is not possible due to the significant bias the dataset has towards younger subjects. 
\vspace{-0.3cm}
\paragraph{Superresolution Losses}
The detail maps generated by FDS \cite{FDS} are of resolution \(4096 \times 4096\) and thus cannot be used directly for training due to GPU-memory constraints. To get around this, we train \(DetP\) on detail maps downsampled to \(256\times 256\). Simultaneously, we finetune a superresolution network, \(RCAN\) \cite{RCAN}, to super-resolve downsampled \(256 \times 256\) patches of \(\mcal{D}(I_{\bf{x}})\) by a factor of 4
\begin{smequation}
    \mcal{L}_{SR}^{RCAN} = L1(RCAN(\mcal{D}(I_{\bf{x}})_{256}^{P}, \mcal{D}(I_{\bf{x}})_{1024}^{P})
\end{smequation}
where \(\mcal{D}(I_{\bf{x}})_{1024}^{P}\) is a randomly sampled patch of resolution \(1024\times 1024 \) from the full-resolution detail map \(\mcal{D}(I_{\bf{x}})\) and  \(\mcal{D}(I_{\bf{x}})_{256}^{P}\) is its downsampled version. During inference, we use \(RCAN\) twice on the predicted detail map \(\mcal{\widetilde{D}}(I_{\bf{y}})\) to upsample it to \(4096 \times 4096\)
\begin{smequation}
    \mcal{\widetilde{D}}(I_{\bf{y}})^{HR} = RCAN(RCAN(\mcal{\widetilde{D}}(I_{\bf{y}})))
\end{smequation}
In the interest of brevity, we will use \(\mcal{\widetilde{D}}(I_{\bf{y}})\) in lieu of \(\mcal{\widetilde{D}}(I_{\bf{y}})^{HR}\) in the remainder of this text.
\vspace{-0.0cm}

\subsection{Rendering Network}
\label{sect:refnet_arch}
\begin{figure*}[t!]
    \centering
    \includegraphics[width=0.72\linewidth]{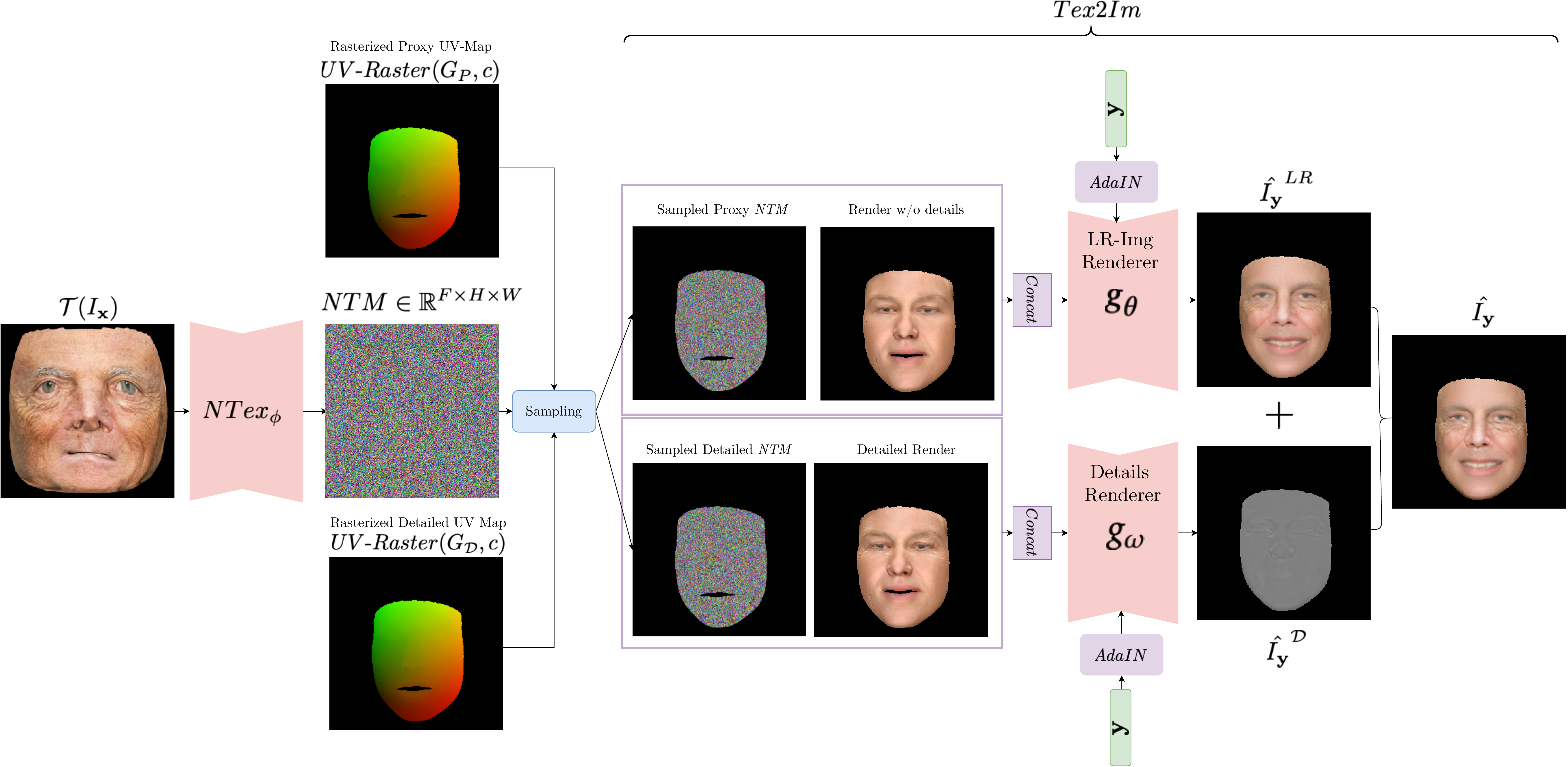}
    \caption{\small{\textbf{The Rendering Network.} The Rendering Network, \(R\) first predicts a Neural Texture Map \(NTM\) from the given input texture map \(\mcal{T}\) using \(NTex_{\phi}\). The \(NTM\) is then rasterized using both the proxy (geometry w/o details) and the detailed geometry and input into an image rendering network \(Tex2Im\). \(Tex2Im\) generates a rendering of the details \(\hat{I_{\mbf{y}}}^{\mcal{D}}\) and a low-resolution image \(\hat{I_{\mbf{y}}}^{LR}\) that contains only detail-invariant image textures. They are added together to generate the final rendered image \(\hat{I_{\mbf{y}}}\).}\vspace{-0.3cm}} 
    \label{fig:rendnet_arch}
\end{figure*}
The rendering network, \(R\), consists of two subnetworks: (1) The Neural Texture prediction network \(NTex_{\phi}\) and (2) The image rendering network \(Tex2Im\).
\vspace{-0.3cm}
\paragraph{Neural Texture Prediction.} \(NTex_{\phi}\) predicts the Neural Textures given  the texture map \(\mcal{T}(I_{\bf{x}})\) of image \(I_{\bf{x}}\):
\begin{smequation}
    NTM =  NTex_{\phi}(\mcal{T}(I_{\bf{x}})); \hspace{0.3cm} NTM \in \mathbb{R}^{F\times H\times W}
\end{smequation}
where \(NTM\) is the predicted neural texture map with \(F\) channels. Unlike in \cite{dnr}, where the \(NTM\) is identity specific, \(NTex_{\phi}\) can be used on any \(\mcal{T}(I_{\bf{x}})\) regardless of identity to generate its corresponding neural texture map. Through training, \(NTex_{\phi}\) learns to extract the appropriate high-dimensional texture features from \(\mcal{T}(I_{\bf{x}})\) such that \(NTM\) can be used to generate a realistic render of the person in \(I_{\bf{x}}\) in any desired expression and view. 
\vspace{-0.3cm}
\paragraph{Image rendering.} The image rendering network, \(Tex2Im\), consists of two branches, the low-res image renderer \(g_{\theta}\) and the detail renderer \(g_{\omega}\). The low-res image renderer generates identity-specific image textures that are invariant to the predicted geometric details, such as the skin-tone or eye color. The detail renderer \(g_{\omega}\) renders the facial geometric details obtained from the detail prediction network, \(DetP\), and adds them to the low-res image generated by \(g_{\theta}\). The separation of the image rendering network into two branches allows each branch to focus on  its respective task and leads to high-quality renderings. 

The low-res image renderer, \(g_{\theta}\)  inputs the $NTM$ sampled using the UV map rasterized by the proxy geometry, \(G_{P}=\{0 \times \widetilde{\mcal{D}}(I_{\mbf{y}}), \alpha_{s}, \hat{\alpha_{e}}\}\), i.e the geometry \textit{without} any details, along with the shaded albedo also rasterized by \(G_{P}\):
\begin{smequation}
    \hat{I_{\bf{y}}^{LR}} = g_{\theta}(\small{\texttt{Sample}}(NTM, \small{UV\shortminus Raster}(G_{P},c)),y, \gamma, l)
\end{smequation}
where \(\gamma\) are the coefficients of the albedo PCA-space of the BFM \cite{gerig2018morphable} and \(l\) are the lighting parameters. Since \(g_{\theta}\) only uses inputs dependent on \(G_{P}\) it generates the image textures that are invariant to details predicted by \(DetP\).

The detail renderer, \(g_{\omega}\) takes as input the $NTM$ sampled using the UV map rasterized by the \textit{detailed} geometry, \(G_{\mcal{D}}=\{\widetilde{\mcal{D}}(I_{\mbf{y}}), \alpha_{s}, \hat{\alpha_{e}}\}\)  along with the shaded albedo also rasterized by \(G_{\mcal{D}}\):
\begin{smequation}
    \hat{I_{\bf{y}}^{\mcal{D}}} = g_{\theta}(\small{\texttt{Sample}}(NTM, \small{UV\shortminus Raster}(G_{\mcal{D}},c)),y, \gamma, l)
\end{smequation}
where \(\gamma\) and \(l\) are the albedo PCA-space and lighting parameters. The final output image is calculated as:
\begin{smequation}
    \hat{I_{\mbf{y}}} = \hat{I_{\bf{y}}^{LR}} + \hat{I_{\bf{y}}^{\mcal{D}}}
\end{smequation}
Since all of the detail invariant textures are already generated by \(g_{\theta}\) in \(\hat{I_{\bf{y}}^{LR}}\), the detail renderer, \(g_{\omega}\) can solely focus on realistically rendering the details predicted by \(DetP\).%

\vspace{-0.3cm}
\subsubsection{Training Losses}
The renderings generated by \(R\) must: (1)  faithfully render the geometric details onto the \RGB space and (2)   be realistic. Neural Rendering \cite{dnr} is designed to address (2) as the high-dimensional neural texture map is able store the fine details of the output texture. Consequently, the details on the rendered image become entirely conditional on the input texture map, \(\mcal{T}(I_{\mbf{x}})\), and ignore the detailed geometry \(G_{\mcal{D}}\). This significantly hurts (1) causing the details on the rendered image to remain unchanged even if the facial geometric details change due to changes in facial expression. In order to ensure the output renderings satisfy (2) we use, along with the branched architecture  discussed in \sect{refnet_arch}, an \textit{Augmented Wrinkle Loss} and the \textit{Detailed Shading Loss} to ensure the geometric details are faithfully rendered onto the output image.    
\vspace{-0.3cm}
\paragraph{Photometric Loss.} The Photometric Loss ensures the rendered images are realistic by re-rendering a  given image \(I_{\bf{x}}\), producing \(\hat{I}_{\bf{x}}\) and comparing it to the ground truth. More specifically, given the texture map \(\mcal{T}(I_{\mbf{x}})\), detailed geometry \(G_{\mcal{D}} = \{\mcal{D}(I_{\mbf{x}}), \alpha_{s}, \alpha_{e}\}\) and action unit \(\bf{x}\), \(I_{\bf{x}}\) is re-rendered using \(R\) as follows
\begin{smequation}
    \begin{split}
        \hat{I}_{\mbf{x}} = R(\mcal{T}(I_{\mbf{x}}), G_{\mcal{D}}, \mbf{x}, c, l, \gamma)
    \end{split}
\end{smequation}
where \(c\) and \(l\) are the camera and lighting parameters of \(I_{\bf{x}}\). The re-rendered image \(\hat{I}_{\mbf{x}}\) is then compared to \(I_{\bf{x}}\)
\begin{smequation}
    \begin{split}
        \mcal{L}_{Photo} &= MSE(\hat{I}_{\mbf{x}}, I_{\bf{x}}) +  L1(\hat{I}_{\mbf{x}}, I_{\bf{x}})\\
        &+  LapLoss(\hat{I}_{\mbf{x}}, I_{\bf{x}}) + PcptL(\hat{I_{\bf{x}}}, I_{\bf{x}})
    \end{split}
\end{smequation}
where \(LapLoss\) is the Laplacian Loss \cite{ling2006diffusion,GLO} and \(PcptL\) is the perceptual loss \cite{perceptual}. In order to ensure the low-res rendering captures the image textures that are invariant to facial details as much as possible, the photometric loss is also applied to the low-res output of \(g_{\theta}\) i.e  \(\hat{I_{\bf{y}}^{LR}}\)
\begin{smequation}
    \begin{split}
        \mcal{L}_{Photo}^{LR} &= MSE(\hat{I_{\bf{x}}^{LR}}, I_{\bf{x}}) +  L1(\hat{I_{\bf{x}}^{LR}}, I_{\bf{x}})\\
        &+ LapLoss(\hat{I_{\bf{x}}^{LR}}, I_{\bf{x}}) + PcptL(\hat{I_{\bf{x}}^{LR}}, I_{\bf{x}})
    \end{split}
\end{smequation}

\begin{figure*}[t!]
    
    \hspace{-0.5cm}\includegraphics[width=1.07\textwidth]{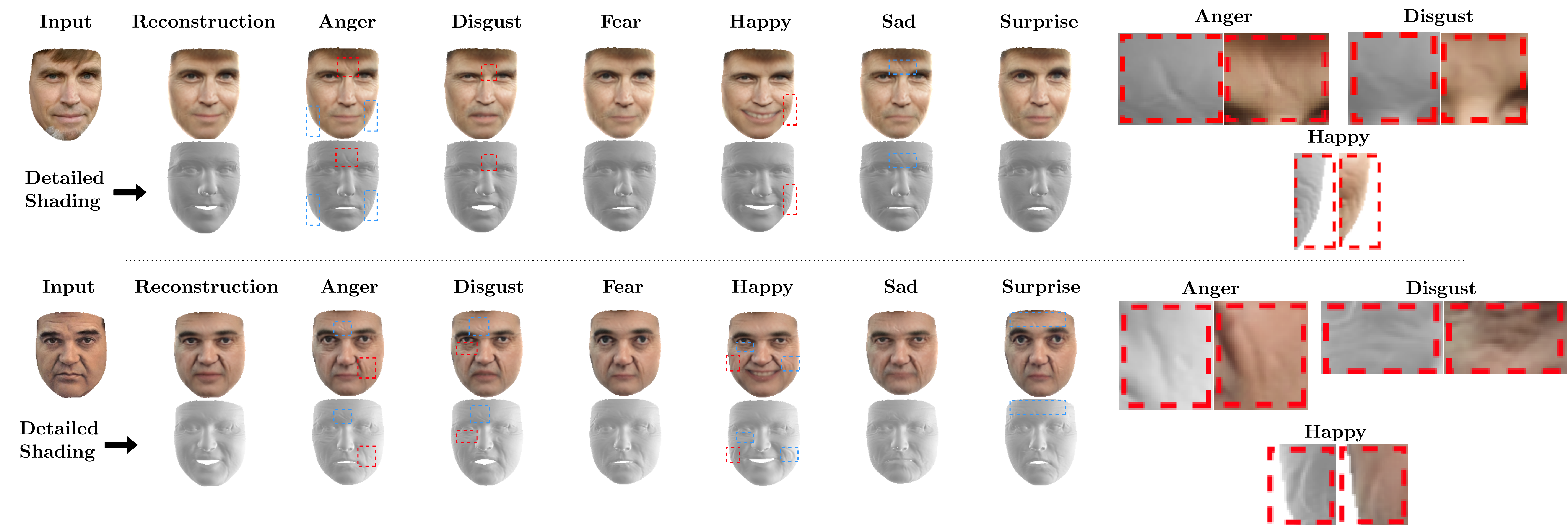}
    \caption{\small{\textbf{Expression Change.} Here we show the results of detail prediction and rendering as the expression changes. The first column is the input image, the second column is the reconstruction and the subsequent columns are the results under different expressions. The first image row  is the output rendering and the second is shading of the detailed geometry. As one can see \(DetP\) is able to generate realistic details depending on the expression being manifested and \(R\) is able to faithfully render them to the image space. We zoom-in on a subset of details  in the final column for greater clarity. \textit{(Please view in high resolution)}}\vspace{-0.5cm}}
    \label{fig:exp_change}
\end{figure*}

\vspace{-0.3cm}
\paragraph{Augmented Wrinkle Loss.} In order to enforce the rendering of geometric details onto the rendered image we add `fake' wrinkles to an image \(I_{\bf{x}}\) and force \(R\) to generate the same. Given the the detailed geometry of \(I_{\bf{x}}\), \(G_{\mcal{D}} = \{\mcal{D}(I_{\mbf{x}}), \alpha_{s}, \alpha_{e}\}\), a geometry with `fake' details \(G_{\mcal{D}}^{*} = \{\mcal{D}(I^{*}_{\mbf{z}}), \alpha_{s}, \alpha_{e}\}\) using the geometric details from some random image \(I^{*}_{\mbf{z}}\)  and the lighting \(l\) of \(I_{\bf{x}}\), the `fake' wrinkles are added as follows
\begin{smequation}
    \begin{split}
        &Shading(I_{\bf_{x}}) = L_{Sph}(G_{\mcal{D}}, l); Shading^{*}(I_{\bf_{x}}) = L_{Sph}(G_{\mcal{D}}^{*}, l)\\
        &I_{\bf_{x}}^{*} = Shading^{*}(I_{\bf_{x}}) \times \left(\frac{I_{\bf_{x}}}{Shading(I_{\bf_{x}})}\right)
    \end{split}
    \label{eq:augw_shade}
\end{smequation}
where \(L_{Sph}\) is the spherical harmonic lighting function and \(l\) are the coefficients of the first 9 spherical harmonics.
The artificially wrinkled image \(I_{\bf_{x}}^{*}\) is now re-rendered using \(R\) 
\begin{smequation}
    \begin{split}
        \hat{I}_{\mbf{x}}^{*} &= R(\mcal{T}(I_{\mbf{x}}), G_{\mcal{D}}^{*}, \mbf{x}, c, l)\\
        \mcal{L}_{AugW} &=  LapLoss(\hat{I}_{\mbf{x}}^{*}, I_{\bf_{x}}^{*})
    \end{split}
\end{smequation}
In order to faithfully reconstruct \(I_{\bf_{x}}^{*}\), \(R\) is forced to rely on the detailed geometry \(G_{\mcal{D}}^{*}\), since the input texture map \(\mcal{T}(I_{\mbf{x}})\), and consequently the neural texture map \textit{contain no} information about the `fake' wrinkles.
\vspace{-0.3cm}
\paragraph{Detailed Shading Loss.} In addition to the Augmented Wrinkle Loss, we also try to predict the shading of the detailed facial geometry from the output rendering \(\hat{I}_{\mbf{x}}\)
\begin{smequation}
    \begin{split}
        &\hat{Shading}(\hat{I}_{\mbf{x}}) = f_{\theta}(\hat{I}_{\mbf{x}})\\
        &\mcal{L}_{DSL} = LapLoss(\hat{Shading}(\hat{I}_{\mbf{x}}), Shading^{*}(I_{\bf_{x}}))
    \end{split}
\end{smequation}
where \(f_{\theta}\) is a small convolutional network (CNN) with only two layers and the shading \(Shading^{*}(I_{\bf_{x}})\) is calculated as in \eq{augw_shade}. We calculate this loss only over the skin region. Since, \(f_{\theta}\) is a small CNN with limited representational capacity, the details must be quite visible on the rendered image \(\hat{I}_{\mbf{x}}\) in order for them to be picked up by \(f_{\theta}\) to generate an accurate shading \(\hat{Shading}(\hat{I}_{\mbf{x}})\).
\vspace{-0.3cm}
\paragraph{Expression Adversarial Loss.} In order to ensure that the rendered output conforms to the target expression we use an expression adversarial loss. Given a rendered image, \(\hat{I}_{\mbf{x}} = R(\mcal{T}(I_{\mbf{x}}), G_{\mcal{D}}, \mbf{x}, c, l)\), manifesting the expression encoded by AU \(\bf{x}\) an expression discriminator, \(\textrm{D}_{\textrm{Exp}}^{\textrm{RGB}}\), outputs 
\begin{smequation}
    \textrm{D}_{\textrm{Exp}}^{\textrm{RGB}}(\hat{I}_{\mbf{x}}) = \{r, \hat{\bf{x}}\} 
\end{smequation}
where \(r\) is a realism score and \(\hat{\bf{x}}\) is the predicted AU. We use the Non-Saturating adversarial loss \cite{goodfellow2014generative} along with the \(R1\) gradient penalty \cite{mescheder2018training} to train \(\textrm{D}_{\textrm{Exp}}\). More specifically, let \(\mcal{P}_{\mcal{I}}\) be the distribution of real images, the adversarial loss can then be written as
\begin{smequation}
    \begin{split}
        \mcal{L}^{\textrm{D}_{\textrm{Exp}}^{\textrm{RGB}}}_{Adv} &= -\mathbb{E}_{\hat{I_{\bf{x}}} \sim R(.)}\left[\mtxtlog \left( 1 \shortminus \textrm{D}_{\textrm{Exp}}^{\textrm{RGB}, \textrm{r}}(\hat{I_{\bf{x}}})\right)\right]\\
                                &- \mathbb{E}_{I_{\bf{x}} \sim \mcal{P}_{\mcal{I}}}\left[\mtxtlog \left(\textrm{D}_{\textrm{Exp}}^{\textrm{RGB}, \textrm{r}}(I_{\bf{x}})\right)\right]\\
                                &+ \mathbb{E}_{I_{\bf{x}} \sim \mcal{P}_{\mcal{I}}}\left[||\nabla \textrm{D}_{\textrm{Exp}}^{\textrm{RGB},\textrm{r}}(I_{\bf{x}})||_{2}^{2}\right]
    \end{split}
\label{eq:adv_ref_gen}
\end{smequation}
where \(\textrm{D}_{\textrm{Exp}}^{\textrm{RGB},\textrm{r}}\) is the realism output head of \(\textrm{D}_{\textrm{Exp}}^{\textrm{RGB}}\). In addition, \(\textrm{D}_{\textrm{Exp}}^{\textrm{RGB}}\) is trained to minimize the  predicted AU error
\begin{smequation}
    \begin{split}
        \mcal{L}^{\textrm{D}_{\textrm{Exp}}^{\textrm{RGB}}}_{Exp} &= \mathbb{E}_{I_{\bf{x}} \sim \mcal{P}_{\mcal{I}}}\left[||[\textrm{D}_{\textrm{Exp}}^{\textrm{RGB},\textrm{AU}}(I_{\bf{x}}) \shortminus \mbf{x}||_{2}^{2}\right]
    \end{split}
\end{smequation}
where \(\textrm{D}_{\textrm{Exp}}^{\textrm{RGB},\textrm{AU}}\) is the AU output head of \(\textrm{D}_{\textrm{Exp}}^{\textrm{RGB}}\). The Rendering Network, \(R\) is trained to minimize  adversarial loss
\begin{smequation}
    \begin{split}
        \mcal{L}^{R}_{ExpAdv} &= \shortminus\mathbb{E}_{I_{\mbf{x}}}\mtxtlog \left(\textrm{D}_{\textrm{Exp}}^{\textrm{RGB},\textrm{r}}(R(.)\right)\\
    \end{split}
\end{smequation}
and  AU loss
\begin{smequation}
    \begin{split}
        \mcal{L}^{R}_{AU} &= \mathbb{E}_{I_{\mbf{x}}}||\textrm{D}_{\textrm{Exp}}^{\textrm{RGB},\textrm{AU}}(R(\cdot)) \shortminus \mbf{y}||_{2}^{2}\\
    \end{split}
\end{smequation}
where \(R(\cdot)\) is to be read as in \eq{refnet_f}.
\vspace{-0.1cm}

\begin{figure}[h!]
    \hspace{-0.5cm}\includegraphics[width=1.05\linewidth]{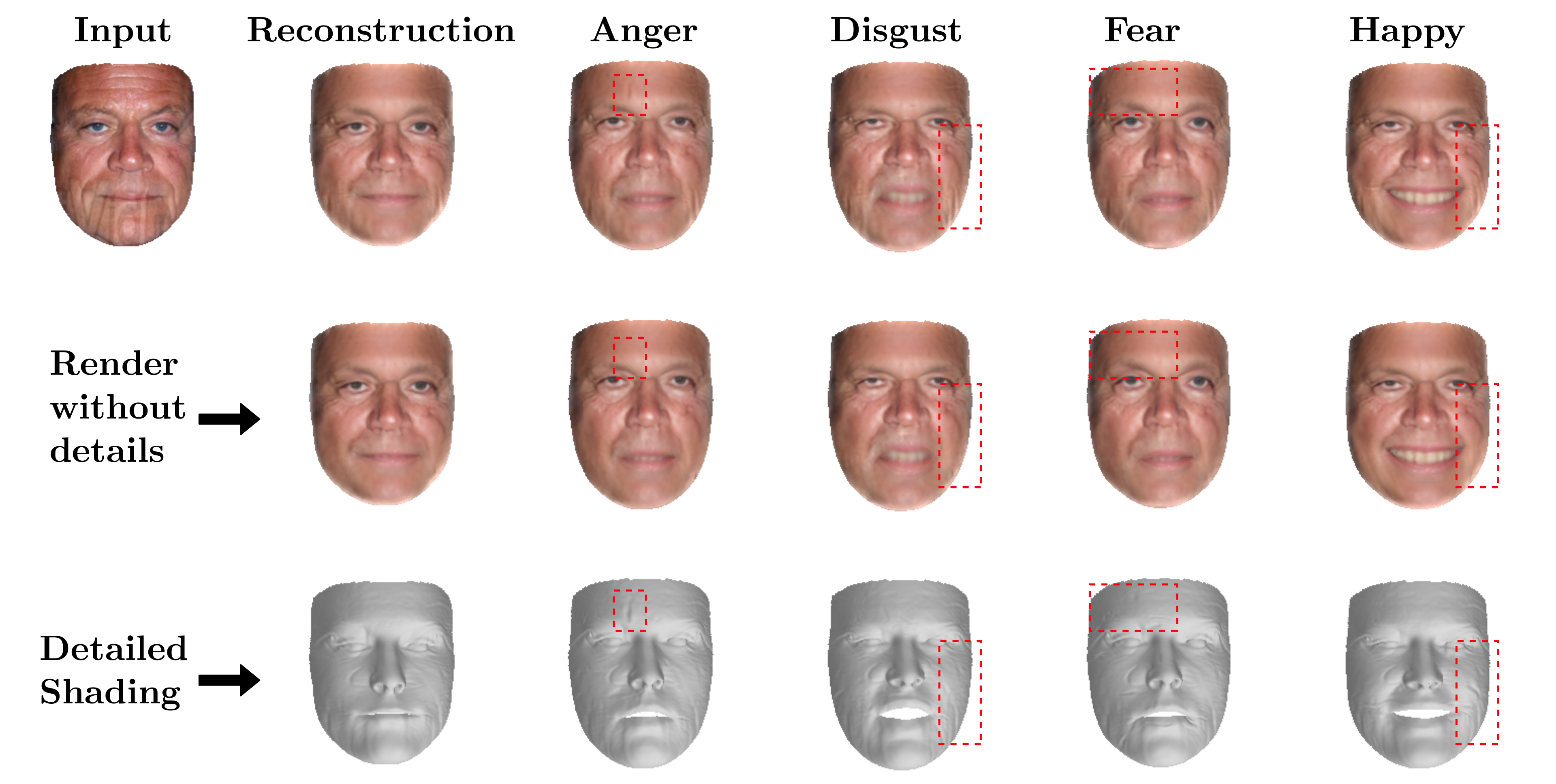}
    \caption{\small{\textbf{Details vs. No Details.} We compare the results of changing the expression both with and without predicting details. \textit{(Please view in high resolution)}}\vspace{-0.5cm}}
    \label{fig:det_nodet}
\end{figure}

\begin{figure*}[h!]
    \centering
    \includegraphics[width=0.9\textwidth]{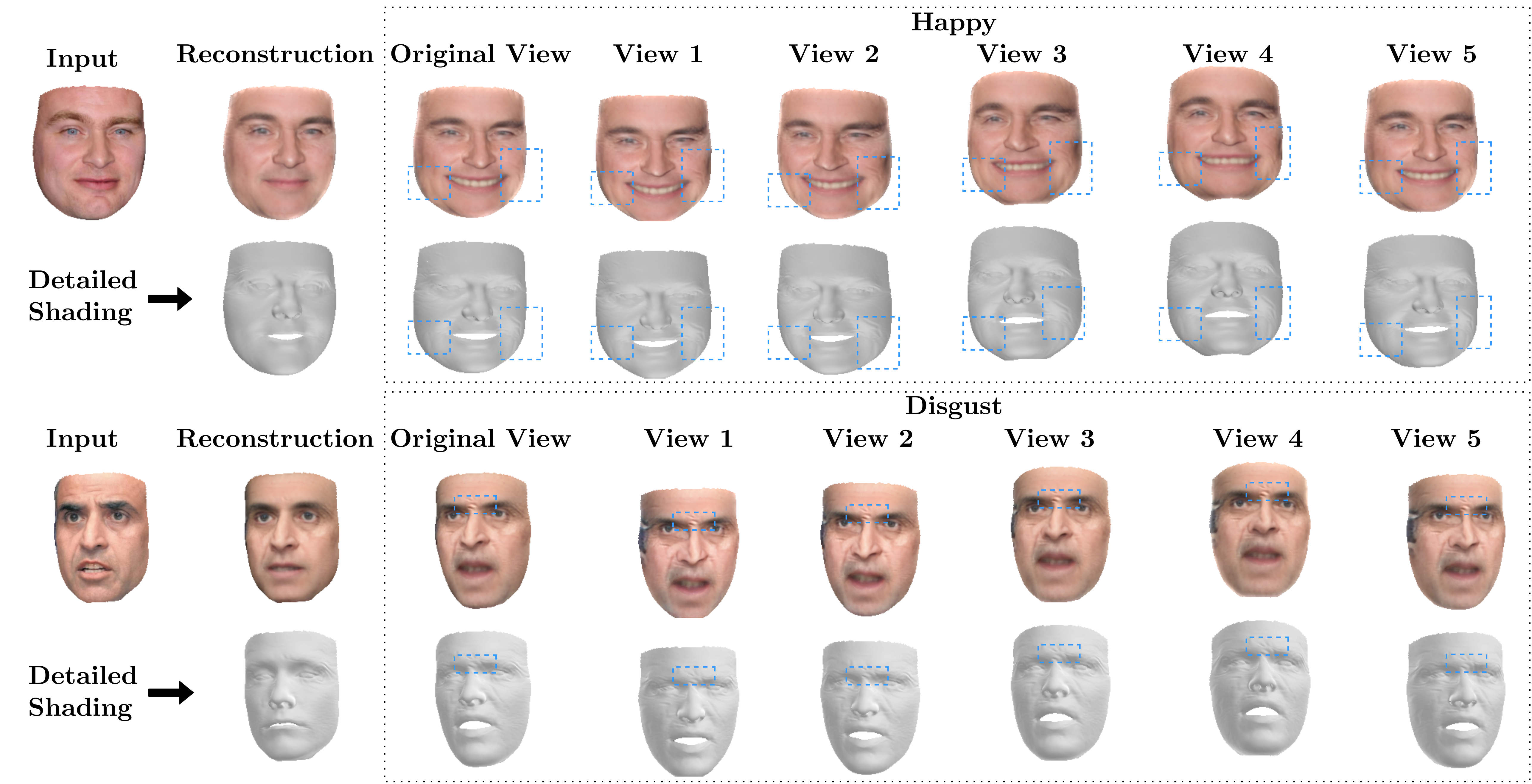}
    \caption{\small{\textbf{View Consistency.} In this figure we demonstrate the consistency of the details rendered by \(R\). A subset of the predicted details from \(DetP\) are marked with blue rectangles. As can be seen, \(R\) renders the details in a consistent manner across views. \textit{(Please view in high resolution)}}\vspace{-0.5cm}}
    \label{fig:view_consistency}
\end{figure*}

\begin{figure}[h!]
    \hspace{-0.5cm}\includegraphics[width=1.1\linewidth]{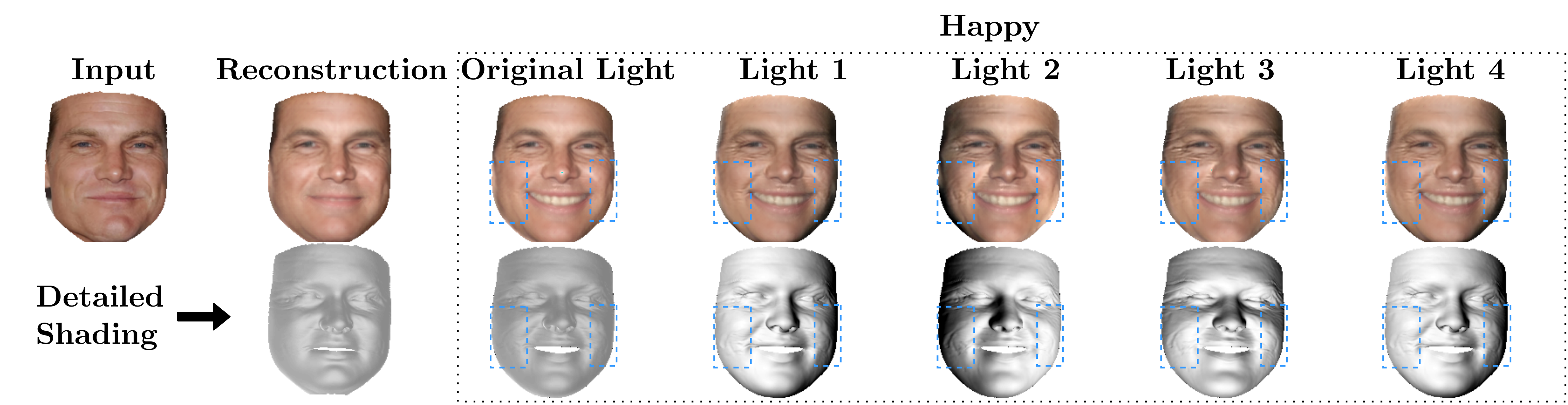}
    \caption{\small{\textbf{Details under varied lighting.} In this figure we show how the appearence of the predicted geometric facial details changes with lighting conditions. \textit{(Please view in high resolution)}}\vspace{-0.3cm}}
    \label{fig:light_change}
\end{figure}

\section{Experiments}
We train the detail prediction network, \(DetP\), and the rendering network \(R\), on 9,000 images from the FFHQ dataset \cite{StyleGAN}. Additionally, 3,000 frames were sampled from the MUG \cite{MUG} and the ADFES \cite{ADFES} datasets to speed-up training of the detail prediction network \(DetP\). Due to memory constraints, \(DetP\) and \(R\) are trained independently. Upon publication we will release the code. %

\vspace{-0.3cm}
\paragraph{Expression change.}  We first demonstrate the  detail prediction and rendering results with expression change.  
\fig{exp_change} shows the result of changing the expression of some input image to a variety of expressions. The first column of \fig{exp_change} is the input image, the second column shows the image reconstructed using the detail map \(\mcal{D}(I_{\mbf{x}})\) predicted by FDS \cite{FDS}. In the subsequent columns \(\mcal{D}(I_{\mbf{x}})\) is used as input to generate \(\mcal{\widetilde{D}}(I_{\mbf{y}})\) where \(\mbf{y} = \{\text{Anger}, \text{Disgust}, \text{Fear}, \text{Happy}, \text{Sad}, \text{Surprise}\}\). The first row shows the final rendered image, i.e the output of \(R\) and the second row shows the shaded geometry with the predicted details, i.e \(G_{\mcal{D}}\). The predicted details and their corresponding rendering are marked out with dashed red and blue rectangles. We zoom in the red rectangles of the last column %
As  seen in the second row of \fig{exp_change}, the details predicted are consistent with the  manifested expression. For example, `Anger' and `Disgust' (3rd and 4th column of \fig{exp_change} respectively) show consistent wrinkling around the forehead and the nose while `Happy' (column 6 \fig{exp_change}) shows consistent wrinkling around the cheeks. Zooming  in the last column of \fig{exp_change} highlights the realism of the predicted details produced by the rendering network, \(R\).

\begin{figure}[t!]
    \hspace{-0.5cm}\includegraphics[width=1.1\linewidth]{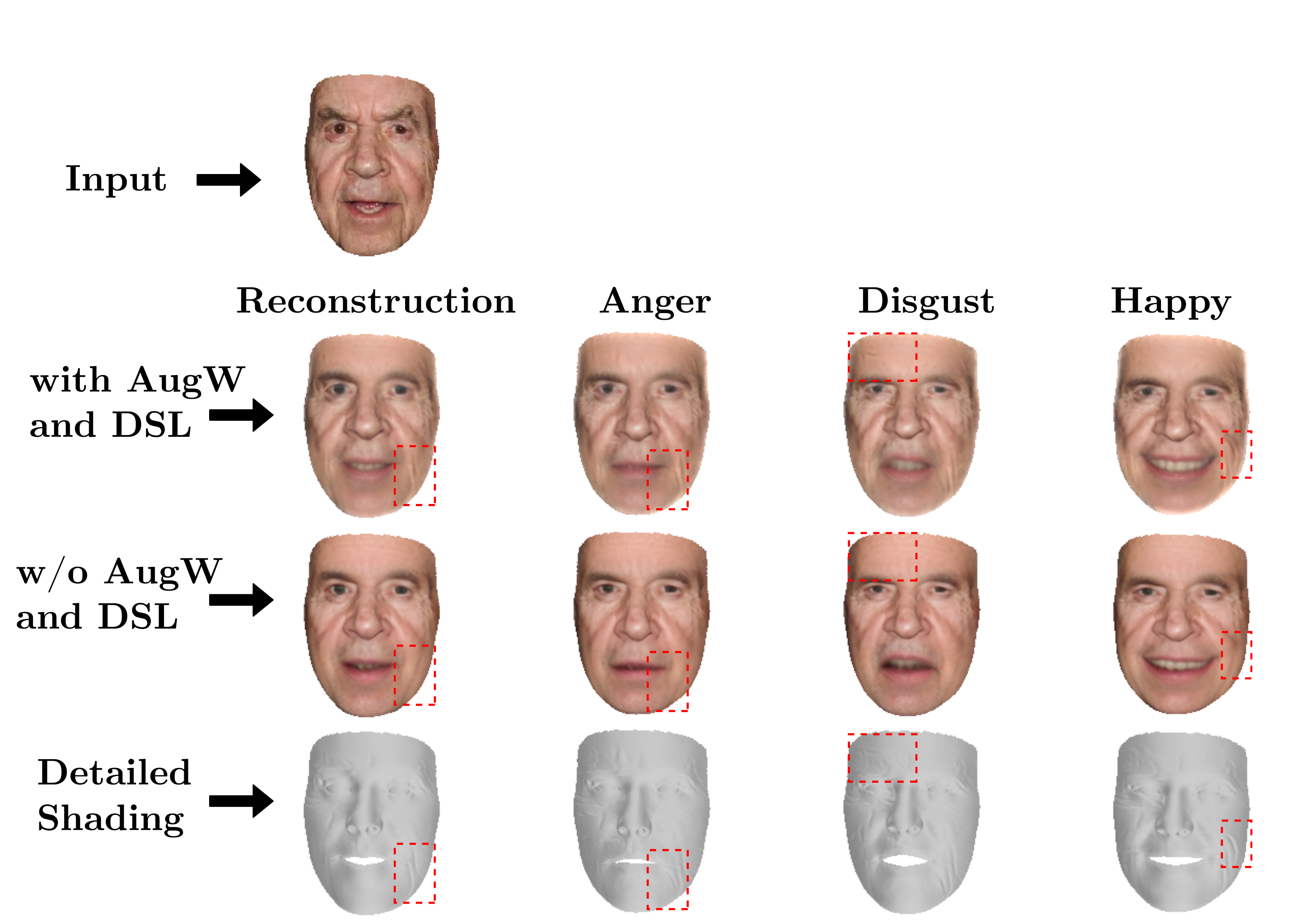}
    \vspace{-5mm}
    \caption{\small{\textbf{Ablating AugW and DSL.} We show that  without \textbf{AugW} and \textbf{DSL} details are not rendered to the image space. \textit{(Please view in high resolution)}}\vspace{-0.7cm}}
    \label{fig:aba_AugW_DP}
\end{figure}

Further, in \fig{det_nodet} we show the utility of predicting details as the expression changes. The first row of \fig{det_nodet} shows the image rendered with the predicted details from \(DetP\) as the expression changes. The second row of images shows the images rendered without details, this is done by setting the predicted detail map to zero. As  seen by comparing the skin appearance marked with the red rectangles, in the first row the skin changes realistically as the expression of the person changes due to the changing facial geometric details, while the skin in the second row remains unchanged, significantly hurting realism. In the supplementary we show further results on expression animation and  encourage the reader to inspect them.
\vspace{-0.5 cm}
\paragraph{View Consistency and lighting variation.}
In \fig{view_consistency} we show the consistency of the details rendered by \(R\) across various views in novel expressions.  While the underlying face model ensures that the detailed geometry is consistent in any view, there is no guarantee its rendering generated by \(R\) would be too. The first column of \fig{view_consistency} is the input image, the second column is the reconstruction of the input in the original expression and view and the third column shows the input image manifesting a novel expression in the view of the input image. The subsequent columns show the input image rendered with the target expression in novel views. The blue rectangles  around the details in the rendered image show they are rendered with high fidelity to the shaded geometry and therefore look consistent across a variety of views. \fig{view_consistency} shows that \(R\) is able to maintain a close to one-to-one correspondence while rendering geometric details to the image space which ensures that they appear consistent across views. In \fig{light_change}, we show how the appearance of the facial geometry details change with changes in lighting. While state-of-the-art lighting manipulation is beyond the scope of this paper (as we do not disentangle lighting), \fig{light_change} demonstrates the utility of predicting facial details on the 3D face geometry to facilitate such a manipulation.
\vspace{-0.5cm}
\paragraph{Ablation Studies.} We  examine the utility of the \textit{Augmented Wrinkle Loss (AugW)} and the \textit{Detailed Shading Loss (DSL)} in rendering the facial geometric details to the image space. \fig{aba_AugW_DP} shows the results of training \(R\) with and without the \textbf{AugW} and \textbf{DP} losses. As  seen by comparing the results in rows 2 and 3 of \fig{aba_AugW_DP}, without those losses \(R\) fails to render the predicted geometric facial details
. 
\vspace{-0.1cm}
\section{Conclusion and Future Work}
\vspace{-0.0cm}
We have presented \MethodName, a method capable of predicting, from a single image, plausible facial geometric details as the expression changes and render them  photo-realistically. Using adversarial losses along with weak supervision we train a detail prediction network capable of predicting plausible facial details for any target expression. A rendering network then renders these details photo-realistically in a manner that is consistent with the target expression and the predicted detailed geometry. The \textit{Augmented Wrinkle Loss} and \textit{Detailed Shading Loss} force the rendering network to use cues from the detailed geometry and not rely solely on the neural texture map to generate the details, ensuring consistency in the appearance of the rendered details across a variety of views.
Due to the lack of explicit disentanglement of the lighting within our method, the renderings we generate are entangled with the input lighting conditions and therefore cannot be controlled independently. The detail prediction network relies on the detail map estimated by FDS \cite{FDS} as input to predict the plausible details of the target expression, therefore it cannot handle occlusions such as glasses or make-up very well as FDS \cite{FDS} fails in those conditions. In future work, we plan to incorporate explicit disentanglement of lighting in order to enable greater control over the final rendering along with explicit modelling occlusions and joint training of \(DetP\) and \(R\).  

\section{Acknowledgements}
This work is supported in part by a Google Daydream Research award, by the Spanish government with  projects HuMoUR TIN2017-90086-R and María de Maeztu Seal of Excellence MDM-2016-0656, by NSF-IUCRC on CVDI, Medpod Inc, by NICHD 1R21 HD93912-01A1, the Partner University Fund and the SUNY2020 Infrastructure Transportation Security Center.

{\small
\bibliographystyle{ieee_fullname}
\bibliography{egbib}

\begin{thebibliography}{10}\itemsep=-1pt

\bibitem{MUG}
Niki Aifanti, Christos Papachristou, and Anastasios Delopoulos.
\newblock The mug facial expression database.
\newblock {\em WIAMIS}, pages 1--4, 2010.

\bibitem{athar2020self}
S Athar, Z Shu, and D Samaras.
\newblock Self-supervised deformation modeling for facial expression editing.
\newblock In {\em IEEE FG}, 2020.

\bibitem{blanz1999morphable}
Volker Blanz, Thomas Vetter, et~al.
\newblock A morphable model for the synthesis of 3d faces.
\newblock In {\em Proc. SIGGRAPH}, 1999.

\bibitem{GLO}
Piotr Bojanowski, Armand Joulin, David Lopez-Pas, and Arthur Szlam.
\newblock Optimizing the latent space of generative networks.
\newblock In {\em Int. Conf. Mach. Learn.}, 2018.

\bibitem{booth20173d}
James Booth, Epameinondas Antonakos, Stylianos Ploumpis, George Trigeorgis,
  Yannis Panagakis, and Stefanos Zafeiriou.
\newblock 3d face morphable models" in-the-wild".
\newblock In {\em IEEE Conf. Comput. Vis. Pattern Recog.} IEEE, 2017.

\bibitem{FDS}
Anpei Chen, Zhang Chen, Guli Zhang, Kenny Mitchell, and Jingyi Yu.
\newblock Photo-realistic facial details synthesis from single image.
\newblock In {\em Int. Conf. Comput. Vis.}, 2019.

\bibitem{StarGAN2018}
Yunjey Choi, Minje Choi, Munyoung Kim, Jung-Woo Ha, Sunghun Kim, and Jaegul
  Choo.
\newblock Stargan: Unified generative adversarial networks for multi-domain
  image-to-image translation.
\newblock In {\em IEEE Conf. Comput. Vis. Pattern Recog.}, 2018.

\bibitem{starganv2}
Yunjey Choi, Youngjung Uh, Jaejun Yoo, and Jung-Woo Ha.
\newblock Stargan v2: Diverse image synthesis for multiple domains.
\newblock In {\em IEEE Conf. Comput. Vis. Pattern Recog.}, 2020.

\bibitem{dou2017end}
Pengfei Dou, Shishir~K Shah, and Ioannis~A Kakadiaris.
\newblock End-to-end 3d face reconstruction with deep neural networks.
\newblock In {\em IEEE Conf. Comput. Vis. Pattern Recog.}, 2017.

\bibitem{FACS}
P. Ekman and W Friesen.
\newblock Facial action coding system: A technique for the measurement of
  facial movement.
\newblock In {\em Consulting Psychologists Press}, 1978.

\bibitem{Genova_2018_CVPR}
Kyle Genova, Forrester Cole, Aaron Maschinot, Aaron Sarna, Daniel Vlasic, and
  William~T. Freeman.
\newblock Unsupervised training for 3d morphable model regression.
\newblock In {\em IEEE Conf. Comput. Vis. Pattern Recog.}, 2018.

\bibitem{gerig2018morphable}
Thomas Gerig, Andreas Morel-Forster, Clemens Blumer, Bernhard Egger, Marcel
  Luthi, Sandro Sch{\"o}nborn, and Thomas Vetter.
\newblock Morphable face models-an open framework.
\newblock In {\em IEEE FG}, 2018.

\bibitem{goodfellow2014generative}
Ian Goodfellow, Jean Pouget-Abadie, Mehdi Mirza, Bing Xu, David Warde-Farley,
  Sherjil Ozair, Aaron Courville, and Yoshua Bengio.
\newblock Generative adversarial nets.
\newblock In {\em Adv. Neural Inform. Process. Syst.}, 2014.

\bibitem{adain}
Xun Huang and Serge Belongie.
\newblock Arbitrary style transfer in real-time with adaptive instance
  normalization.
\newblock In {\em Int. Conf. Comput. Vis.}, 2017.

\bibitem{pix2pix2016}
Phillip Isola, Jun-Yan Zhu, Tinghui Zhou, and Alexei~A Efros.
\newblock Image-to-image translation with conditional adversarial networks.
\newblock In {\em IEEE Conf. Comput. Vis. Pattern Recog.}, 2017.

\bibitem{jackson2017large}
Aaron~S Jackson, Adrian Bulat, Vasileios Argyriou, and Georgios Tzimiropoulos.
\newblock Large pose 3d face reconstruction from a single image via direct
  volumetric cnn regression.
\newblock In {\em Int. Conf. Comput. Vis.}, 2017.

\bibitem{perceptual}
Justin Johnson, Alexandre Alahi, and Li Fei-Fei.
\newblock Perceptual losses for real-time style transfer and super-resolution.
\newblock In {\em Eur. Conf. Comput. Vis.}, 2016.

\bibitem{StyleGAN}
Tero Karras, Samuli Laine, and Timo Aila.
\newblock A style-based generator architecture for generative adversarial
  networks.
\newblock In {\em IEEE Conf. Comput. Vis. Pattern Recog.}, 2019.

\bibitem{Kim2018DeepVP}
H. Kim, P. Garrido, A. Tewari, Weipeng Xu, Justus Thies, M. Nie{\ss}ner,
  Patrick P{\'e}rez, C. Richardt, M. Zollh{\"o}fer, and C. Theobalt.
\newblock Deep video portraits.
\newblock {\em ACM Transactions on Graphics (TOG)}, 37:1 -- 14, 2018.

\bibitem{kim2018inversefacenet}
Hyeongwoo Kim, Michael Zollh{\"o}fer, Ayush Tewari, Justus Thies, Christian
  Richardt, and Christian Theobalt.
\newblock Inversefacenet: Deep monocular inverse face rendering.
\newblock In {\em IEEE Conf. Comput. Vis. Pattern Recog.}, 2018.

\bibitem{ling2006diffusion}
Haibin Ling and Kazunori Okada.
\newblock Diffusion distance for histogram comparison.
\newblock In {\em IEEE Conf. Comput. Vis. Pattern Recog.}, 2006.

\bibitem{mescheder2018training}
Lars Mescheder, Andreas Geiger, and Sebastian Nowozin.
\newblock Which training methods for gans do actually converge?
\newblock In {\em Int. Conf. Mach. Learn.}, 2018.

\bibitem{Nagano2018paGANRA}
K. Nagano, Jaewoo Seo, J. Xing, Lingyu Wei, Zimo Li, S. Saito, Aviral Agarwal,
  Jens Fursund, and H. Li.
\newblock pagan: real-time avatars using dynamic textures.
\newblock {\em ACM Trans. Graph.}, 2018.

\bibitem{pumarola2020ganimation}
Albert Pumarola, Antonio Agudo, Aleix~M Martinez, Alberto Sanfeliu, and
  Francesc Moreno-Noguer.
\newblock Ganimation: One-shot anatomically consistent facial animation.
\newblock {\em International Journal of Computer Vision}, 128(3):698--713,
  2020.

\bibitem{schroff2015facenet}
Florian Schroff, Dmitry Kalenichenko, and James Philbin.
\newblock Facenet: A unified embedding for face recognition and clustering.
\newblock In {\em IEEE Conf. Comput. Vis. Pattern Recog.}, 2015.

\bibitem{sela2017unrestricted}
Matan Sela, Elad Richardson, and Ron Kimmel.
\newblock Unrestricted facial geometry reconstruction using image-to-image
  translation.
\newblock In {\em Int. Conf. Comput. Vis.}, 2017.

\bibitem{NeuralFace2017}
Z. Shu, E. Yumer, S. Hadap, K. Sunkavalli, E. Shechtman, and D. Samaras.
\newblock Neural face editing with intrinsic image disentangling.
\newblock In {\em IEEE Conf. Comput. Vis. Pattern Recog.}, 2017.

\bibitem{tewari2019fml}
Ayush Tewari, Florian Bernard, Pablo Garrido, Gaurav Bharaj, Mohamed Elgharib,
  Hans-Peter Seidel, Patrick P{\'e}rez, Michael Z{\"o}llhofer, and Christian
  Theobalt.
\newblock Fml: Face model learning from videos.
\newblock In {\em Proceedings of the IEEE Conference on Computer Vision and
  Pattern Recognition}, pages 10812--10822, 2019.

\bibitem{tewari2017self}
Ayush Tewari, Michael Zollh{\"o}fer, Pablo Garrido, Florian Bernard, Hyeongwoo
  Kim, Patrick P{\'e}rez, and Christian Theobalt.
\newblock Self-supervised multi-level face model learning for monocular
  reconstruction at over 250 hz.
\newblock In {\em IEEE Conf. Comput. Vis. Pattern Recog.}, 2018.

\bibitem{Thies2020NeuralVP}
Justus Thies, M. Elgharib, A. Tewari, C. Theobalt, and M. Nie{\ss}ner.
\newblock Neural voice puppetry: Audio-driven facial reenactment.
\newblock In {\em ECCV}, 2020.

\bibitem{dnr}
Justus Thies, M. Zollh{\"o}fer, and M. Nie{\ss}ner.
\newblock Deferred neural rendering.
\newblock {\em ACM Transactions on Graphics (TOG)}, 2019.

\bibitem{tran2019towards}
Luan Tran, Feng Liu, and Xiaoming Liu.
\newblock Towards high-fidelity nonlinear 3d face morphable model.
\newblock In {\em In Proceeding of IEEE Computer Vision and Pattern
  Recognition}, Long Beach, CA, June 2019.

\bibitem{tran2018nonlinear}
Luan Tran and Xiaoming Liu.
\newblock Nonlinear 3d face morphable model.
\newblock In {\em IEEE Computer Vision and Pattern Recognition (CVPR)}, Salt
  Lake City, UT, June 2018.

\bibitem{extreme3D}
Anh Tuấn~Trần, Tal Hassner, Iacopo Masi, Eran Paz, Yuval Nirkin, and
  G{\'e}rard Medioni.
\newblock Extreme 3d face reconstruction: Seeing through occlusions.
\newblock In {\em IEEE Conf. Comput. Vis. Pattern Recog.}, 2018.

\bibitem{ADFES}
Job Van Der~Schalk, Skyler~T Hawk, Agneta~H Fischer, and Bertjan Doosje.
\newblock Moving faces, looking places: validation of the amsterdam dynamic
  facial expression set (adfes).
\newblock {\em Emotion}, 11(4):907, 2011.

\bibitem{zakharov2019few}
Egor Zakharov, Aliaksandra Shysheya, Egor Burkov, and Victor Lempitsky.
\newblock Few-shot adversarial learning of realistic neural talking head
  models.
\newblock In {\em Int. Conf. Comput. Vis.}, 2019.

\bibitem{RCAN}
Yulun Zhang, Kunpeng Li, Kai Li, Lichen Wang, Bineng Zhong, and Yun Fu.
\newblock Image super-resolution using very deep residual channel attention
  networks.
\newblock In {\em Eur. Conf. Comput. Vis.}, 2018.

\bibitem{CycleGAN2017}
Jun-Yan Zhu, Taesung Park, Phillip Isola, and Alexei~A Efros.
\newblock Unpaired image-to-image translation using cycle-consistent
  adversarial networks.
\newblock In {\em Int. Conf. Comput. Vis.}, 2017.

\bibitem{reda}
Wenbin Zhu, HsiangTao Wu, Zeyu Chen, Noranart Vesdapunt, and Baoyuan Wang.
\newblock Reda:reinforced differentiable attribute for 3d face reconstruction.
\newblock In {\em Proceedings of the IEEE/CVF Conference on Computer Vision and
  Pattern Recognition (CVPR)}, June 2020.

\end{thebibliography}
}

\end{document}